# Big Meaning: Qualitative Analysis on Large Bodies of Data Using AI

*Samuel Flanders[1], Melati Nungsari[1], Mark Cheong Wing Loong[2]*

[1] *Asia School of Business, Kuala Lumpur, Malaysia*
[2] *Malaysia School of Pharmacy, Monash University*

**Abstract**

This study introduces a framework that leverages AI-generated descriptive codes to indicate a text's *fecundity*—the density of unique human-generated codes—in thematic analysis. Rather than replacing human interpretation, AI-generated codes guide the selection of texts likely to yield richer qualitative insights. Using a dataset of 2,530 Malaysian news articles on refugee attitudes, we compare AI-selected documents to randomly chosen ones by having three human coders independently derive codes. The results demonstrate that AI-selected texts exhibit approximately twice the fecundity. Our findings support the use of AI-generated codes as an effective proxy for identifying documents with a high potential for meaning-making in thematic analysis.

## Introduction

Recent research has shown capacity for LLM powered coding methodologies to generate codes for thematic analysis (TA) (Dai, Xiong, and Ku, 2023; Morgan, 2023; De Paoli, 2024). However, researchers may not want to delegate such an important task to a computer, and may consider the process of reading and analyzing a corpus to be an essential part of TA–something lost even if the AI generated codes are of acceptable quality by some measure. In this paper we develop an alternative application for these AI-generated codes: they can serve as an accurate signal of a text's *fecundity*, which we define as the expected density of unique human generated codes.[1]

We build a framework for selecting a subset of documents–the corpus to be read–from a larger superset of documents. This method maximizes the density of unique AI-generated codes of the corpus, which we hypothesize will also yield high fecundity. To test this hypothesis, we utilize a dataset of 2530 Malaysian news articles on the topic of refugees, all of which have been AI-coded in relation to the research question "What attitudes towards refugees are present in Malaysian news media?" From this dataset, we create a test corpus containing both AI-selected and randomly-selected articles, ordered randomly[2], and ask three human coders to develop their own codes for the corpus. We find a large, statistically significant difference in the average fecundity–density of unique codes across all three coders–of AI-selected and randomly-selected articles. Fecundity is approximately twice as high in the AI-selected documents.

---

[1] Per character of text in our implementation, though other metrics could reasonably be used.
[2] AI-selected and randomly-selected articles are themselves interleaved randomly so the coders cannot identify distinct corpora while reading.

In this paper we do not conceptualize the AI-generated codes as measures–or even noisy measures–of the "true" codes in the corpus. Keeping with the norms of approaches such as TA and grounded theory, we conceive of the final codes as the result of an interaction between a coder and a text (aka researcher-derived codes), often describing implicit meanings of the data that are identified through the coder's knowledge of the subject matter. Conversely, AI-generated codes are better described as being data-derived or "descriptive" codes that closely describe the explicit or semantic content of the data. However, we argue that some documents are more fecund than others, denser with text that could be made meaningful for a particular research question and that captures distinct issues rather than being highly redundant with other members of the corpus. We further argue that, with an appropriate prompt capturing that research question, AI-generated codes can serve as a useful proxy for that fecundity, identifying documents that are likely to be more useful to researchers.

This paper is premised on the fecundity as a useful operationalization of a text's capacity for meaning-making, but also the idea that, when performing TA, a larger set of unique codes are better than a smaller one when attempting to identify and develop themes that capture important concepts from the data analysed, at least up to a point (Saunders et al., 2018). We do not commit to the controversial idea that saturation–however defined–is a necessary component of TA, but aver that broad engagement with a corpus is necessary and that it is possible to have too few codes and miss key facets of the phenomenon under study.

That being said, various operationalizations of the concept of saturation are often used in TA, so we provide some results on how AI-selected corpora perform with respect to various notions of saturation. We find impressive results for AI-selected corpora–coders achieve saturation far faster than with randomly selected articles in metrics where saturation is possible. We also show that, in many cases, standard saturation methodologies may suggest saturation has occurred when, with respect to more fecund corpora such as our AI-selected corpora, very few of the codes that human coders will develop have yet been encountered.

## Literature Review

This paper applies AI tools to propose a systematic, structured mechanism to analyze qualitative datasets, taking a richer view of the deployment of research methods to extract meaning and findings from datasets than traditional viewpoints. Specifically, it takes a stance, in line with Paley (2000), that quantitative and qualitative research methods are simply tools and not contradictory philosophical paradigms. Hence, applying new AI technology to enhance and increase the capacity of researchers to analyze more bodies of text and larger datasets is a "pareto improvement" – i.e., we employ complementarity of these methods to overcome weaknesses that each may have (dos Santos, 1999). While quantitative research focuses on numbers and measures, stemming from a belief that more is better, and qualitative research focuses more on meaning, qualitative research can still benefit from methods that would allow for larger (in quantity) amounts of data to be analyzed and interpreted to develop theory and meaning.

*Thematic Analysis*

There are many approaches to TA, often dramatically different from one another in method, purpose, and philosophical underpinnings. Positivist approaches like Coding Reliability (Boyatzis, 1998) focus on codes as simple 'domain summaries' and convert qualitative information into quantitative data to be statistically processed. This approach can admit direct AI coding as a substitute for human coders, though our approach can still be relevant for researchers using this family of methods if they're unsatisfied with the quality of the AI codes. Other approaches include Reflexive TA (Braun and Clarke, 2006) and Critical Realist TA (Fryer, 2022). While they differ from one another in their epistemological underpinnings and the degree to which TA can generate objective knowledge about the world, these approaches insist that the researcher cannot be removed from the process of research–that the knowledge to be created is not fully latent in the text, but is a result of interplay between text and researcher. Researchers using these approaches may find our methods particularly useful, as we focus on finding fecund texts without committing to the particular AI codes generated.

*AI and Thematic Analysis*

Another relevant strand of literature is the nascent use of LLMs for TA and allied methods of textual analysis. More traditional machine learning techniques have been applied to qualitative analysis in the past. For example: grounded theory (Nelson, 2020), qualitative coding and content analysis (Renz et al., 2018), and TA itself (Gauthier & Wallace, 2022). However, these methods are much more limited when dealing with the meaning of a text, only able to measure narrow concepts like sentiment and relatively flat constructs like word frequencies and topics constructed via Latent Dirichlet Allocation. They can coarsely categorize textual data, but generally can't understand it in the way an LLM can. Similarly, A variety of computational techniques have been used to select rich or diverse documents such as Maximal Marginal Relevance (Carbonell and Goldstein, 1998), and various notions of entropy, e.g. Moradi et al. (1998). Our paper differs from these more traditional methods in that the metric it uses to identify rich documents focuses directly on the research question and the intermediate research goal–developing codes, rather than more broad and coarse natural language processing metrics.

However, since the release of ChatGPT in late 2022, LLMs have frequently been used for thematic analysis. "Within months, established qualitative software packages started integrating GPT-based features: for example, ATLAS.ti added an AI auto-coding assistant in March 2023, MAXQDA in April 2023, and NVivo by late 2024" (Silver and Paulus, 2025). Morgan (2023) found that AI was able to reproduce many of the original themes, but was less successful at locating subtle, interpretive themes, and more successful at reproducing concrete, descriptive themes.

Dai, Xiong, and Ku (2023) propose a framework where large language models (LLMs) like GPT-3.5 collaborate with human coders to conduct thematic analysis, reducing the labor and time traditionally required for qualitative data analysis. This framework integrates LLMs in generating initial codes and refining them through iterative discussions with human coders, maintaining coding quality comparable to two human coders while still benefiting from the efficiency of AI-driven initial coding. Similarly, Chew et al. (2023) present a methodology that uses LLMs for deductive coding, focusing on co-developing codebooks, testing validity, and using LLMs to automate coding tasks, reducing the burdens associated with traditional content analysis. In contrast, De Paoli (2024) investigates whether GPT-3.5 can conduct inductive thematic analysis on interview datasets, attempting to execute the reflexive coding process of Braun and Clarke (2006) with LLM integration. Turobov et al. (2024) provides another iterative framework for joint human-AI coding.

Most relevant to our paper, De Paoli and Mathis (2024) explore how inductive thematic saturation (ITS) can be used as a metric to validate the initial coding phase of TA performed by large language models (LLMs). While they do not address corpus selection, their focus on LLM coding and saturation is closely aligned with this paper. The authors propose a method to measure ITS by tracking the ratio of unique codes to total codes generated by an LLM over time. A challenge for this paper is that saturation is usually framed in terms of reaching a point where new codes are no longer discovered, but they never reach such a point in their datasets and must instead measure when the rate of novel over repeat codes stabilizes–often at a very high level, implying conventional notions of saturation are not achieved. Additionally, they do not utilize bootstrap methods to characterize convergence to saturation, making their results difficult to interpret, as heterogeneity across documents adds overwhelming noise to their measures of saturation convergence.

## Methods

### *Data Description*

The full set of documents was drawn from a dataset built to study attitudes towards refugees in Malaysia compiled from Malaysian news articles available online. To choose sources, we performed an "incognito" google search for "malaysia news" and excluded promoted results, with a target of five distinct sources to ensure variety. Many search results were unusable for various reasons shown in Table A1, so our five sources come from the top 11 results.

To collect the data, we performed the query '"malaysia" "refugees"' for each source's website search and collected all articles from January 1st, 2017 to the date of search, which varied from March to May 2023. To generate the passages to be coded by AI, we split the articles by line break and dropped all passages less than 100 characters long. We also generated one paragraph summaries of each article using GPT in order to provide context to GPT when it analyzed the passages. Figure A2 shows the distribution of article lengths and paragraph counts for the dataset.

Four samples were selected from this dataset: a treatment sample using AI selection and a randomly selected control for Round 1 (see AI Coding) and treatment and control samples for Round 2. Table 1 shows summary statistics on document length for these four samples and the full dataset, as well as predicted fecundity for the four samples.[3]

*Table 1: Summary of Predicted Fecundity (AI code density) and Text Length (characters) for the four samples from the dataset*

| Condition | Variable | Mean | N | 95% CI Lower | 95% CI Upper | 25th Percentile | 75th Percentile |
| --- | --- | --- | --- | --- | --- | --- | --- |
| Round 1 - Randomly Selected | AIcodedensity | 3.31 | 10 | 2.13 | 4.49 | 2.64 | 4.56 |
| Round 1 - Randomly Selected | textlen | 3,185 | 10 | 1,755 | 4,614 | 1,857 | 3,977 |
| Round 1 - AI Selected | AIcodedensity | 2.99 | 40 | 2.54 | 3.45 | 1.97 | 3.97 |
| Round 1 - AI Selected | textlen | 1,968 | 40 | 1,786 | 2,150 | 1,621 | 2,106 |
| Round 2 - Randomly Selected | AIcodedensity | 2.20 | 18 | 1.36 | 3.03 | 0.73 | 3.83 |
| Round 2 - Randomly Selected | textlen | 4,225 | 18 | 3,339 | 5,110 | 3,503 | 5,438 |
| Round 2 - AI Selected | AIcodedensity | 5.64 | 34 | 5.54 | 5.75 | 5.50 | 5.85 |
| Round 2 - AI Selected | textlen | 2,080 | 34 | 1,871 | 2,289 | 1,624 | 2,465 |
| Full Dataset | textlen | 3,963 | 2,530 | 3,866 | 4,059 | 2,237 | 5,000 |

### *AI Coding*

We generated two sets of AI codes over the course of this project, which we denote Round 1 and Round 2. Only Round 2 is utilized in the analysis, as Round 1 codes suffered from a high rate of false positives, making them unusable as a predictor of fecundity. In Round 1, to generate AI codes, we prompted a LLM, GPT-3.5 Turbo (Henceforth GPT), with the following prompt:

```
'Read a passage from a news article summarized here: ### ' + str(summary) + '
### passage: ### ' + str(excerpt)+ ' ### In 12 words or less, give the theme of
this specific passage as it embodies, relates to or reflects attitudes towards
refugees in Malaysia, or return "Irrelevant"'
```

Where *summary* was a GPT generated summary of the full article and excerpt was the *paragraph* to be coded.

We found numerous shortcomings in this approach and devised a more robust coding process, summarized in Figure 1[4]. Essentially, we identified and removed clusters of irrelevant passages before coding to avoid false positives, developed a more structured prompting process to ensure the generated codes were more detailed, and broke the dataset into 400 semantically related clusters, generating 4 high quality codes for each cluster and providing those codes to GPT as a form of dynamic few-shot prompting, giving the LLM four examples of good codes for related passages to use as reference. This process generated 17,549 unique AI codes. More detail is provided in Appendix B.

---

[3] Fecundity measures unique codes, and uniqueness depends on the set of codes being compared. In this case, it measures uniqueness amongst the codes in the samples and is not equivalent to a measure of fecundity over the whole dataset.
[4] The number of articles in this figure is greater than the 2,530 quoted in Data Description as we dropped extremely short articles from the sampling frame.

*Figure 1: Round 2 Coding Diagram*

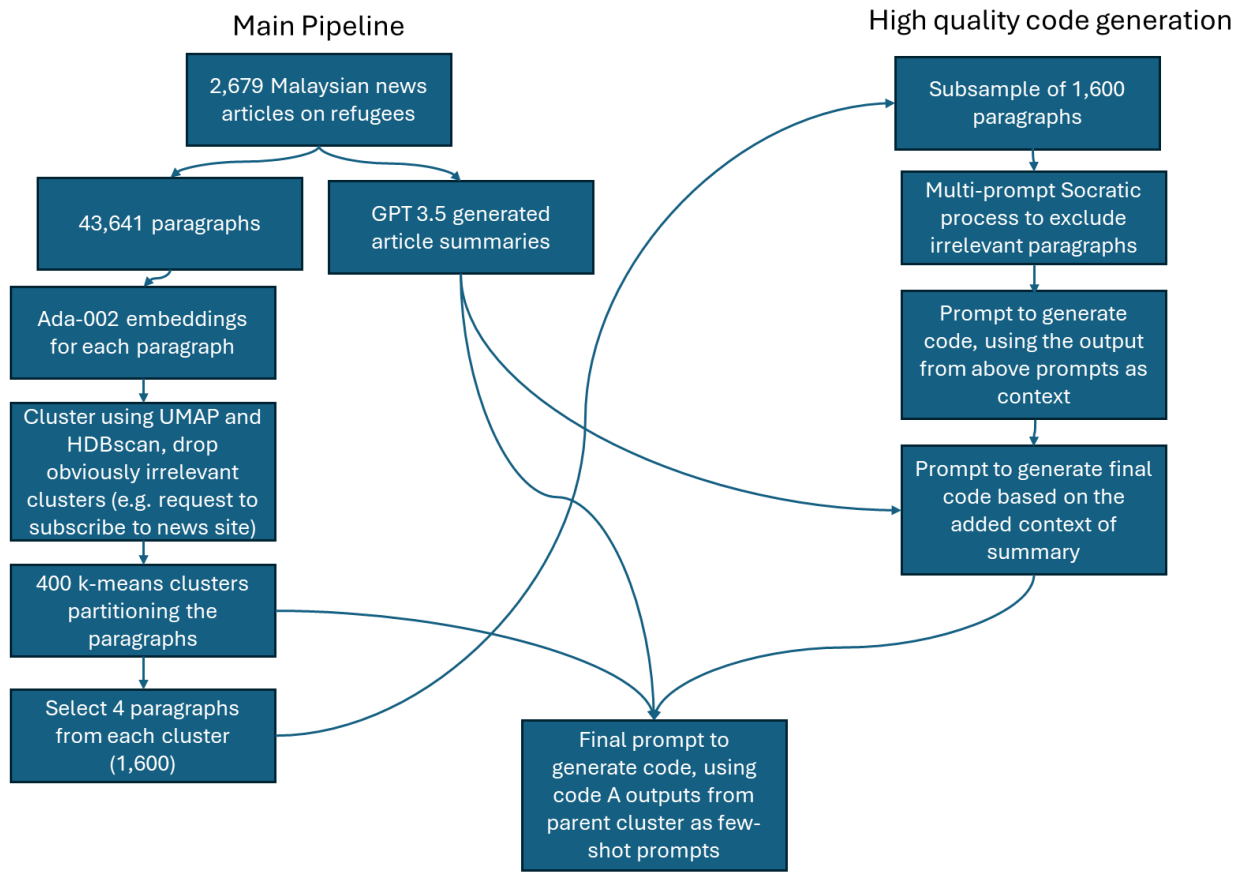

***Article Selection***

Given a set of articles, their predicted number of codes, and their length, an obvious way to select articles would be to sort articles by expected code count divided by length (predicted code density) and pick the top n articles for some desired number n. However, some codes are much more common than others, and we base our analysis on the assumption that a diversity of codes is more useful than numerous documents covering the same issues. We could weight each code by its inverse frequency in the larger set of documents, but in this methodology the human readers will never read that whole set of documents, only the selected articles–a code could be very frequent in the overall dataset and thus selected against by this method could nonetheless be absent in the selected corpus when clearly it should have been selected. We want a method that will only start selecting against codes once they're well represented in the corpus.

Therefore, our approach is instead to find a corpus of a given size that maximizes code density while discounting multiple copies of the same code. To accomplish this, we need to do discrete submodular optimization, which we execute in the Apricot python package. Also, rather than choosing a fixed number of documents, which would make the total size of the selected corpus unpredictable since documents vary in length, we set a maximum total character length for the corpus, which we set as the average length of a corpus on n documents. Formally, taking S as the set of all documents to be considered for inclusion, we choose a corpus $A \subset S$ to maximize

$$\sum_{i \in C} \sqrt{\sum_{a \in A} f_{i,a}}$$ subject to the constraint that $\sum_{a \in A} l_a < L$

$l_a$ is the length of document $a$ in characters and L is a threshold corpus size in characters.

$C$ is the set of all AI generated codes found in the documents in S.

$f_{i,a}$ is the frequency across the entire corpus of the $ith$ in document $a$.

Our value function depends on the square root of the number copies of each code, ensuring that variety is valued rather than just selecting the most frequent codes.[5] Because we are generating a corpus with the average length of n documents, the resulting number of documents may be larger if the selected documents are shorter than average or smaller if the selected documents are longer than average. In our case, the optimization model selected articles that were shorter than average and we selected 34 documents given a length constraint that the corpus be less than or equal to 20 average documents.

*Experimental Design*

Our goal is to assess our methodology's ability to select a high fecundity corpus and our approach is to create a treatment and control group of documents and have 3 expert human coders code both, then compare the results. To do this, we select two tranches of documents: one control tranche randomly from the full set of 2530 and the treatment tranche using AI-generated codes and the optimization method above. We then randomize the order of the documents and create a docx file with all the selected news articles. This randomization is intended to ensure that coders cannot distinguish between AI-coded and random sections, which could bias their codes, and that there is no bias in the ordering of the tranches as the character and frequency of human codes might change over the course of the coding exercise.

---

[5] Other options could include taking the log of the frequency or even maximizing the number of unique codes. We found that the square root generated A.i. estimated fecundity scored more correlated with human code density than either of these options, likely because a particular A.I. code does not match one-to-one with human code choices, so 10 copies of one A.I. code may correspond to several distinct human codes.

Our primary outcome for the experiment was set as the number of unique codes per 1000 characters. We measured this based on the entire corpus–random and AI-selected–for every article. One complication is that, if we are counting unique codes across a corpus, that means each code can only be counted once–or with a weight of one–even if it appears in multiple copies of the code. To address this, we reweight each code by the inverse of its frequency, such that if there are n copies of the code, each has a weight of 1/n and their total contribution to the unique count is 1. Formally, for each document D, we calculate $unique_D = \sum_{i \in D} 1/f_i$, where $f_i$ is the frequency of the i*th* code of document D across the entire corpus S. Finally, We calculate $fecundity_D = \frac{unique_D}{length_D} * 1000$ and compare the fecundity values for the treatment and control arms.

We must note that this paper focuses on the *second* experiment in this project. The initial attempt used an earlier version of the AI-coding model and the AI-selected articles did no better than random in terms of fecundity. However, this initial attempt provided us with a dataset of human-coded passages that we could test later AI-coding models against. The current, more sophisticated model appeared much more capable of predicting human code density, so we created a new AI-selected corpus using the new coding and repeated the experiment. The coders and methodology remained the same across experiments.

### *Human Coding*

Human coders were provided with eight randomly selected news articles each and instructed to follow a two-phase structured coding process. Before beginning the coding process, coders went through a familiarization step where they read through the content of each article multiple times, noting relevant phrases and passages. During this familiarization phase, the human coders were also encouraged to make memos of ideas, patterns, codes and themes that emerged during their reading. In the first coding phase, they engaged in iterative open coding where they independently identified and applied codes to relevant text segments while maintaining a code glossary to ensure conceptual consistency. During the second phase, coders critically reviewed and refined their codes, ensuring relevance and accuracy. Coded text could be decoded, re-coded with additional codes, or had their codes revised. In the final stage, the human coders identified major themes on the attitudes towards refugees as depicted in the media by analyzing the coded data through pattern identification and conceptual clustering.

## Analysis

### *Main Results*

Table 2 shows the main results of the paper. The mean number of unique human generated codes per 1000 characters is about 1.4 in the randomly selected articles, and, across model specifications, we find the increase in the code rate for the AI-selected articles to be 1.2-1.5, approximately doubling the rate at which codes are found in a document, and the results are highly statistically significant in every specification.

Specification 1 includes only the documents coded in the second experiment. There are three articles that were selected by both the first experiment's AI selection procedure and the second round's select procedure. Since they had already been coded by the same coders, they were exempted from the second round of coding, and are excluded here as the two coding sessions happened two months apart and the coders' coding behavior could have changed over that interval, changing the measured fecundity.

Specification 2 includes those three documents, using the codes found in the first round.

Specification 3 uses the same sample as 1 but includes order controls–the index of each article and its square–to address any issues generated by asymmetric distribution of treatment and control articles across the corpus. Coders were asked to read the whole corpus before coding and had already completed a coding exercise in the first experiment, which should have helped avoid changes in coding behavior across the corpus, but we still see a decline in the frequency of codes from the beginning to the end–possible coding fatigue. However, including order controls only increases the estimated benefit of AI-selection.

To address the concern that the coders may have been coding at a different rate in the first round of coding, Specification 4 includes both the 3 overlap AI-selected articles and all the randomly selected articles from the first round, and includes a round dummy to control for any average differences in coding behavior across rounds. It seems that the coders may have coded slightly less vigorously in round 2, though the effect is not statistically significant.

Specification 5 combines the sample of Specification 4 with the order controls.

Specification 6 uses the same sample as Specification 1, but weights observations according to article length under the reasoning that the outcome of interest is not the average article's fecundity, but the average fecundity across articles, in which each article's fecundity must be weighted by its length.

*Table 2: The effect of AI selection of fecundity*

|  | \multicolumn{6}{c}{Dependent variable: Fecundity} |  |  |  |  |  |
| --- | --- | --- | --- | --- | --- | --- |
|  | (1) | (2) | (3) | (4) | (5) | (6) |
| Constant | 1.382*** | 1.309*** | 14.405*** | 2.415*** | 1.810* | 1.191*** |
|  | (0.353) | (0.355) | (7.125) | (0.866) | (0.985) | (0.249) |
| AI-selected | 1.282*** | 1.401*** | 1.449*** | 1.283*** | 1.269*** | 1.410*** |
|                          | (0.439) | (0.439)      | (0.437) | (0.426)      | (0.424) | (0.366) |
| Round                    |         |              |         | -0.517       | -0.881  |         |
|                          |         |              |         | (0.507)      | (0.984) |         |
| Index                    |         |              | -0.328* |              | 0.055   |         |
|                          |         |              | (0.189) |              | (0.041) |         |
| Ind^2                    |         |              | 0.002   |              | -0.000  |         |
| Observations             | 48      | 52           | 48      | 62           | 62      | 48      |
| R²                       | 0.157   | 0.169        | 0.237   | 0.133        | 0.172   | 0.244   |
| Adjusted R²              | 0.138   | 0.153        | 0.185   | 0.104        | 0.114   | 0.228   |
| Residual Std. Error      | 1.454 (df=46) | 1.505 (df=50) | 1.414 (df=44) | 1.556 (df=59) | 1.547 (df=57) | 1.459 (df=46) |
| F Statistic              | 8.539*** (df=1; 46) | 10.203*** (df=1; 50) | 4.566*** (df=3; 44) | 4.538** (df=2; 59) | 2.959** (df=4; 57) | 14.875*** (df=1; 46) |
| Overlap selected Articles | False  | True         | False   | True         | True    | False   |
| Old random articles      | False   | False        | False   | True         | True    | False   |
| Article order controls   | False   | False        | True    | False        | True    | False   |
| Weighted by article length | False | False        | False   | False        | False   | True    |

Note: *p<0.1; **p<0.05; ***p<0.01

*Fecundity and Article Length*

As seen in Figure 2, the AI-selected articles tend to be shorter than the randomly-selected articles. Indeed, Figure 3 shows that, among the randomly selected documents that were coded (both rounds), shorter documents tend to have higher fecundity. We'd like to be able to say that this is simply because shorter news articles in this dataset tend to be more dense with information and longer articles more padded and plodding, and that our method has detected that. However, there is an obvious alternative explanation for this: coders may experience fatigue coding longer articles and become less diligent, leading to fewer recorded codes. That the AI model predicts the same could then be just a coincidence that has nothing to do with fecundity as we've defined it. Additionally, we'd like to be able to say that the AI can predict fecundity better than document length.

To test these hypotheses, we run a few analyses. If the hypothesis is true–if predicted fecundity is only predicting human codes via its correlation with text length–controlling for text length should render the AI predicted fecundity insignificant as a predictor of human code density and AI predicted fecundity should add little to the $R^2$. In Table 3, Specification 1 provides a baseline, predicting human code density with text length using the round 2 data, and finds a strong negative relationship between text length and code density–longer documents are less fecund. However, Specification 2 shows that AI code density explains much more of the variation in human code density than text length alone–the $R^2$ rises dramatically–and the coefficient on text length shrinks to nearly zero–there's little association between human fecundity and variations in text length not predicted by AI fecundity. This suggests that AI code density is not simply acting as a noisy proxy for text length when predicting human code density. In Specification 3 we include the overlap and randomly-selected documents from round 1 along with a round dummy and find an effect of text length that is not just small but positive, although it is far from statistically significant.

*Figure 2:*

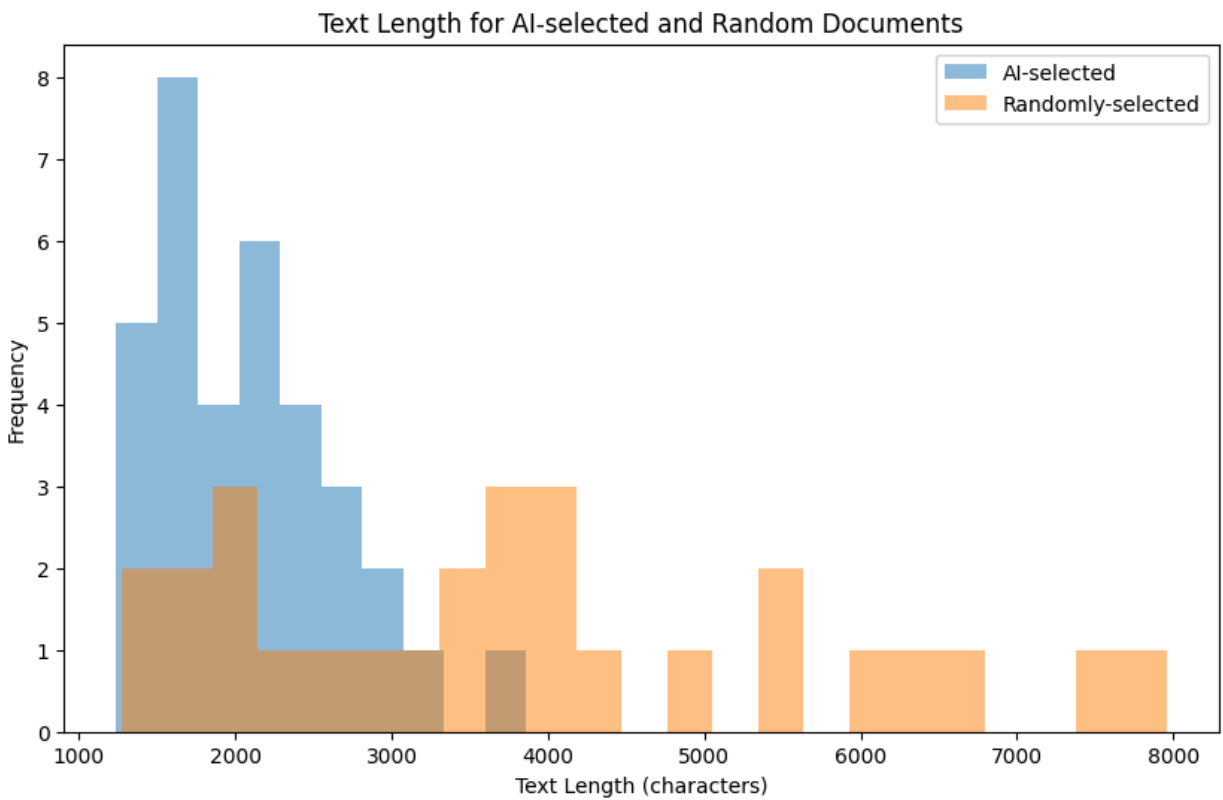

As another robustness check, in Table 4 we predict text length using AI code density and add the residuals from this regression–the variation in text length that cannot be explained by AI code density–to Table 2 Specification 5. If AI code density were merely a proxy for text length then, in the cases where AI code density fails to predict text length, The deviation of text length from AI code density should itself predict lower fecundity and lower the estimated effect of AI code density. However, we find a (non-significant) positive sign on the length residual coefficient and a slightly higher estimate for the treatment effect. In the first stage prediction of text length, AI code density predicts less than half ($R^2 = 0.453$) of variation in text length, so the insignificance of the residual is not a mechanical effect of lower variation in the residual vis-a-vis the component of text length predicted by AI code density–the opposite is true.

Finally, we look at the distribution of comments in the articles themselves. Figure 4 shows the relative position of the median code comment (0 for beginning, 1 for end) graphed against document length. If coders are decreasing effort as they move through long documents, the codes should bunch at the beginning and we should see the median comment position[6] for codes moving earlier in the document for longer documents–the moving average curve in Figure 4 should slope downward. However, we see no clear relationship between length and median comment position.

*Figure 3:*

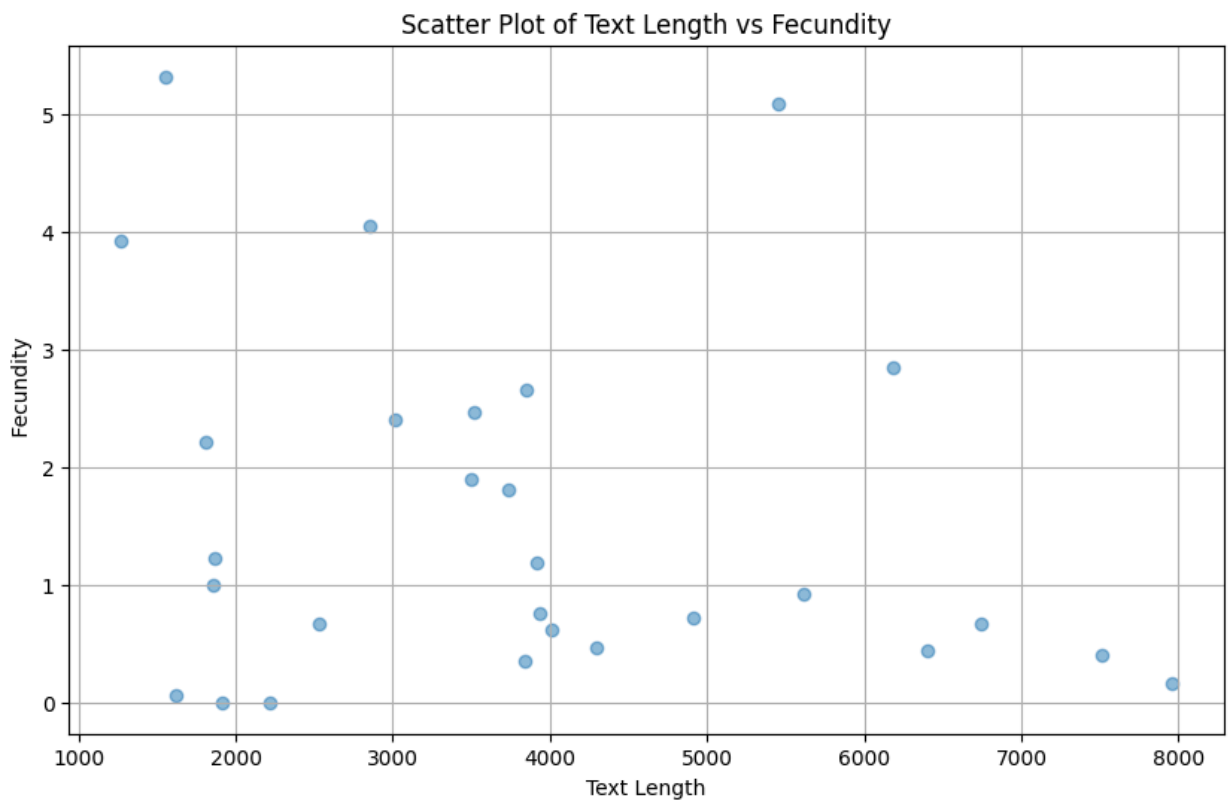

*Fecundity is unique human codes per 1000 characters, the scatterplot shows all articles from the random selection control group across both rounds of coding.*

*Table 3: Fecundity and Document Length*

|  | Dependent variable: Fecundity | | |
|---|---|---|---|
|  | (1) | (2) | (3) |

---

[6] We measure a code's position by the average of the beginning and end position (in characters from the beginning of the document) of the paragraph that code's comment is embedded in in the docx coding file.

|  |  |  |  |
|---|---|---|---|
| Predicted Fecundity |  | 363.674** | 493.727*** |
|  |  | (148.489) | (130.432) |
| Round |  |  | -295.213 |
|  |  |  | (440.814) |
| Constant | 3.472*** | 0.893 | 0.469 |
|  | (0.433) | (1.130) | (1.099) |
| Text Length (000's) | -0.437*** | -0.100 | 0.052 |
|  | (0.132) | (0.186) | (0.155) |
| Observations | 48 | 48 | 62 |
| $R^2$ | 0.192 | 0.287 | 0.290 |
| Adjusted $R^2$ | 0.175 | 0.256 | 0.254 |
| Residual Std. Error | 1.423 (df=46) | 1.351 (df=45) | 1.420 (df=58) |
| F Statistic | 10.951*** (df=1; 46) | 9.070*** (df=2; 45) | 7.914*** (df=3; 58) |

Note: $^{*}p<0.1$; $^{**}p<0.05$; $^{***}p<0.01$

Table 4: The effect of AI selection of fecundity with Length Residuals

|  | Dependent variable: Fecundity |
|---|---|
| Round | -0.887 |
|  | (0.988) |
| Constant | 1.839* |
|  | (0.990) |
| index | 0.056 |
|  | (0.041) |
| ind^2 | -0.000 |
|  | (0.000) |
| AI-Selected | 1.296*** |
|  | (0.427) |
| Length Residuals | 0.125 |
|  | (0.174) |
| Observations | 62 |
| $R^2$ | 0.180 |
| Adjusted $R^2$ | 0.106 |
| Residual Std. Error | 1.554 (df=56) |
| F Statistic | 2.451** (df=5; 56) |

Note: $^{*}p<0.1$; $^{**}p<0.05$; $^{***}p<0.01$

*Figure 4:*

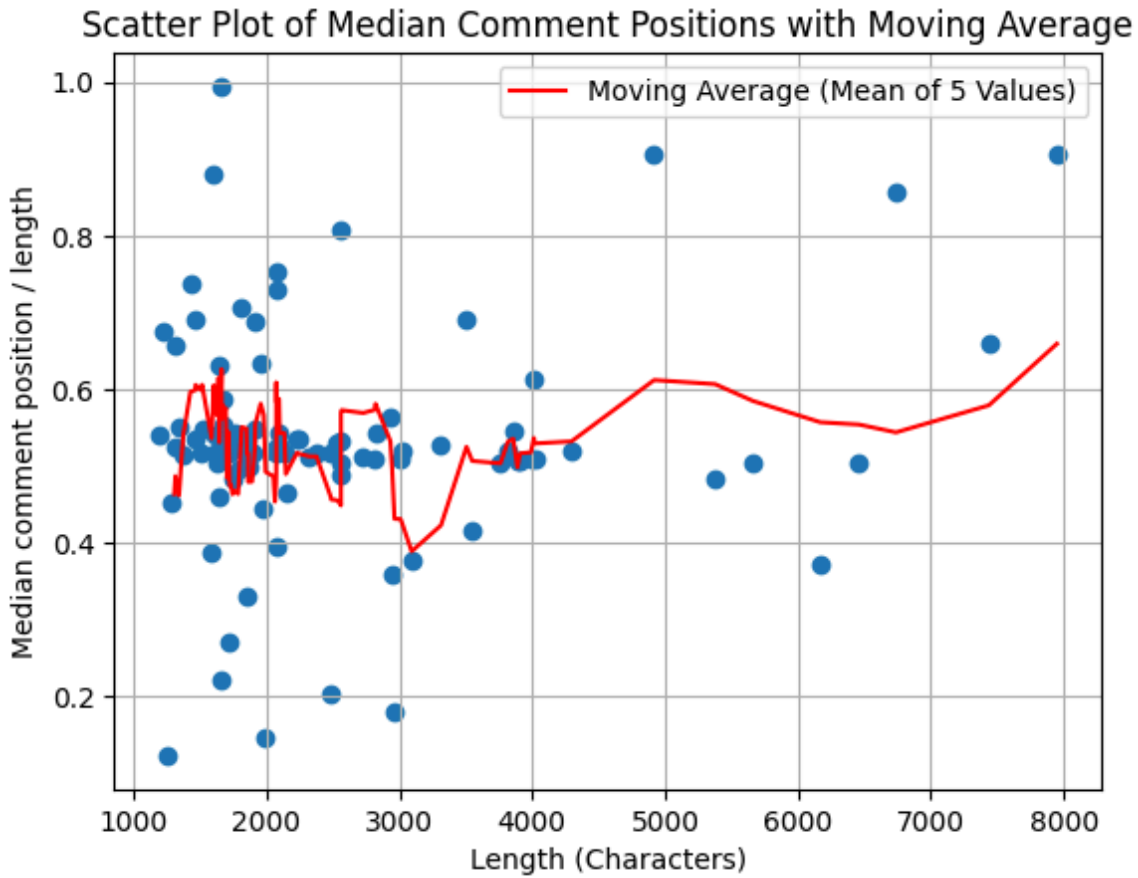

### *Repeated Codes and Saturation*

A question every researcher using TA must answer is "how much is enough?" Whether we borrow the theoretical saturation approach from grounded theory and look holistically at when we know enough about a phenomenon to build a theory or define it as information redundancy and rely on simple rules of thumb like coding 12 documents, a choice must be made to end coding. However, there is wide disagreement on how such a decision should be made. Our results allow us to shed new light on this issue, especially with respect to more quantitative approaches to measuring saturation. We specifically focus on code saturation, high-frequency code saturation, and theme saturation.

We'll first consider the most straightforward method of measuring saturation: counting unique codes. Figure 5 shows the cumulative number of unique human generated codes for both the AI-selected documents and the randomly-selected documents from round 2. To achieve saturation we'd like to see a decline in–and ideally a cession of–new codes as we code more documents. If you squint you can see the curves sloping downward, but in this corpus we get nowhere near saturation so defined. But perhaps this is an artifact of the particular order of the

documents–cumulative unique code graphs are very noisy, as their dynamics are governed by two factors: 1) variation in the density of codes from document to document, and 2) a trend towards fewer of the current document's codes being unique as we move from left to right and more unique codes have already been accumulated. In practice, 1) often dominates 2) and the shape of such a graph often says more about the order of documents than the long term trend in new code discovery. A simple way around this is bootstrapping[7]. We can randomize the order of the documents and generate the cumulative unique codes repeatedly. Averaging the results will give us a smooth curve that eliminates the kinks from 1) by distributing high and low code density documents evenly over the entire cumulative process, leaving only the average increase in codes from document to document and 2). Further, we can use these simulations to generate 95% confidence intervals[8] for the graphs showing how far the cumulative process might deviate from the average for a particular ordering of texts.

*Figure 5*

---

[7] Our approach is similar to Guest, Namey, and McKenna (2016), but we use sampling without replacement to avoid downward bias in the unique code counts and compensate with a Finite Population Correction. See Appendix C

[8] While our confidence intervals account for heterogeneity in document code density, they do not account for path dependence in coding. We should expect a coder's work to depend not just on the passage in front of them, but also on their previous experience, which depends on the order of documents. We cannot model this, and the upshot is that the confidence intervals are likely too narrow as they miss this source of variation in coding across different document orders. The fact that our coders had already done a coding exercise on the same topic and were instructed to read all documents before coding should help here, but is no guarantee that path dependence is not relevant.

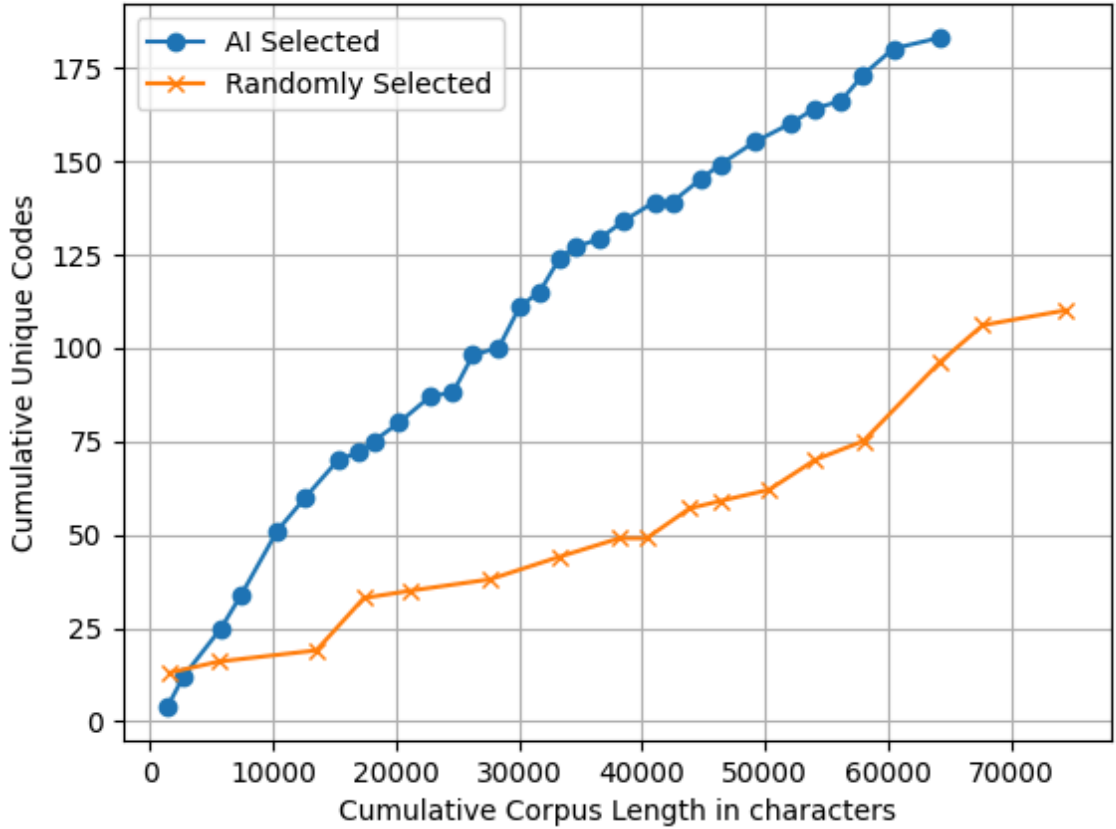

Figure 6 presents these results for our AI-selected and random documents. The gentle downward curves caused by factor 2) are now unmistakeable, but so are the continued upward trajectories. Given these curves, it seems unlikely that we'd run out of codes if we doubled or tripled the corpus size. It should be noted that our coding process generated very granular, open-ended codes. With such a coding process, we should not generally expect to achieve saturation in terms of unique codes, but TA methodologies with coarser codes may plausibly reach a flattening of this curve.

*Figure 6:*

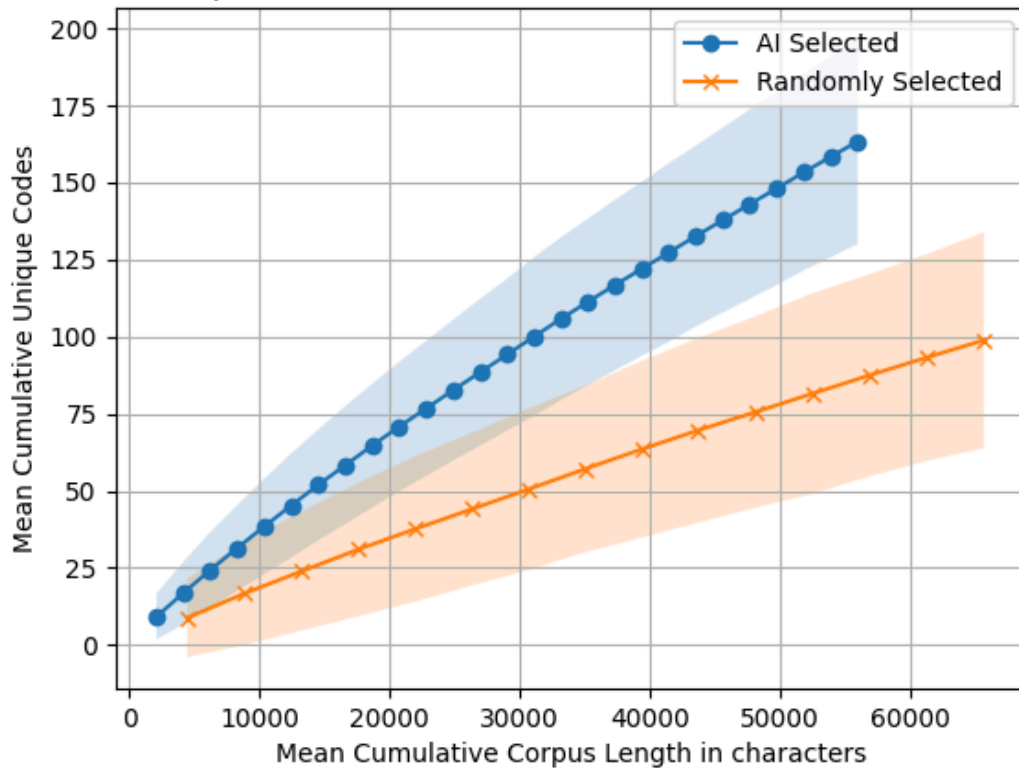

If unique codes cannot tell us we've read enough, what about focusing on common codes? Many authors, like Guest, Bunce, and Johnson (2006), Francis et al. (2010), and Hennick, et al. (2017) measure saturation of "shared" or "high-frequency" codes, and this measure is attractive as the codebook tends to converge faster when evaluated on those grounds.[9]

*Figure 7:*

---

[9] It is also attractive as a measure of "important" codes, winnowing rare and therefore unimportant codes, though one can argue that rare codes can also be important to a researcher's understanding of a phenomenon.

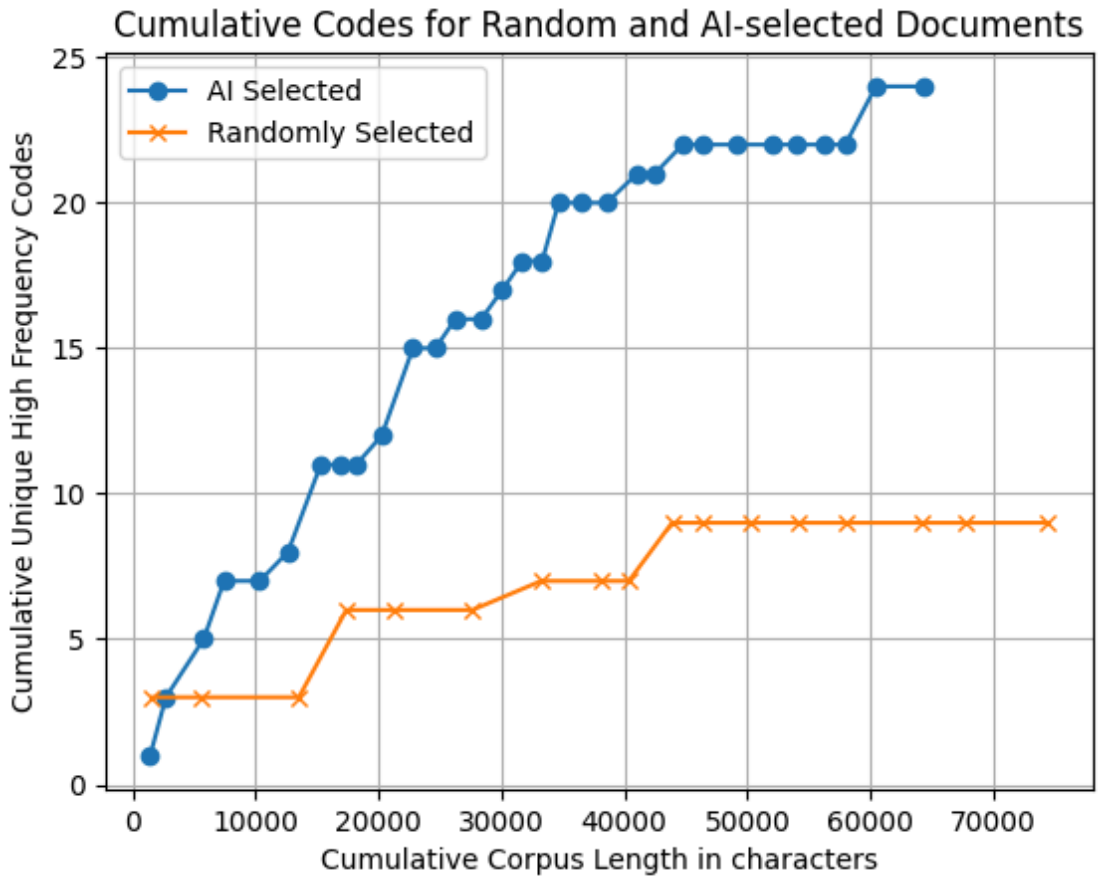

We now present the same analysis as in Figures 3 and 4 looking exclusively at high-frequency codes, which we define here as codes that appear 3 or more times in the full codebook[10]. Figure 7 shows the actual results from our document order and Figure 8 the smoothed results from the bootstrap of random document orders. What we find should be concerning. In Figure 7, we see what appears to be saturation according to the widely used "10 + 3" stopping rule of Francis et al. (2010)–that is, a minimum of 13 documents and 3 consecutive documents without new codes–for both the AI-selected and randomly-selected corpora. However, the number of unique high-frequency codes is vastly different: 9 for random-selection and 22[11] for AI-selection. This demonstrates the strength of the AI-selection methodology, but also the inadequacy of this method of quantifying saturation–clearly the apparent saturation in the randomly selected documents does not correspond to any meaningful way of information redundancy. In fact, this approach is fundamentally prone to identifying saturation regardless of whether it exists or not: imagine each document could contain only one copy of a code. Then, when evaluating saturation based on codes that occur n or more times in a corpus, the last n-1 documents must show no new high frequency codes by construction as the first instance of a code with n or more instances must occur on the nth to last document or earlier. In reality, a document can contain multiple copies of the same code, but the later documents in the ordering will still show an artificial flattening of new high frequency codes due to this dynamic.

---

[10] The codebook of codes from second round AI-selected documents for the AI-selected documents and the codebook of codes from second round randomly-selected documents for the randomly-selected documents.

[11] The graph reaches 24 but the first run of 3 consecutive documents without new codes occurs at 22 unique high frequency codes.

*Figure 8:*

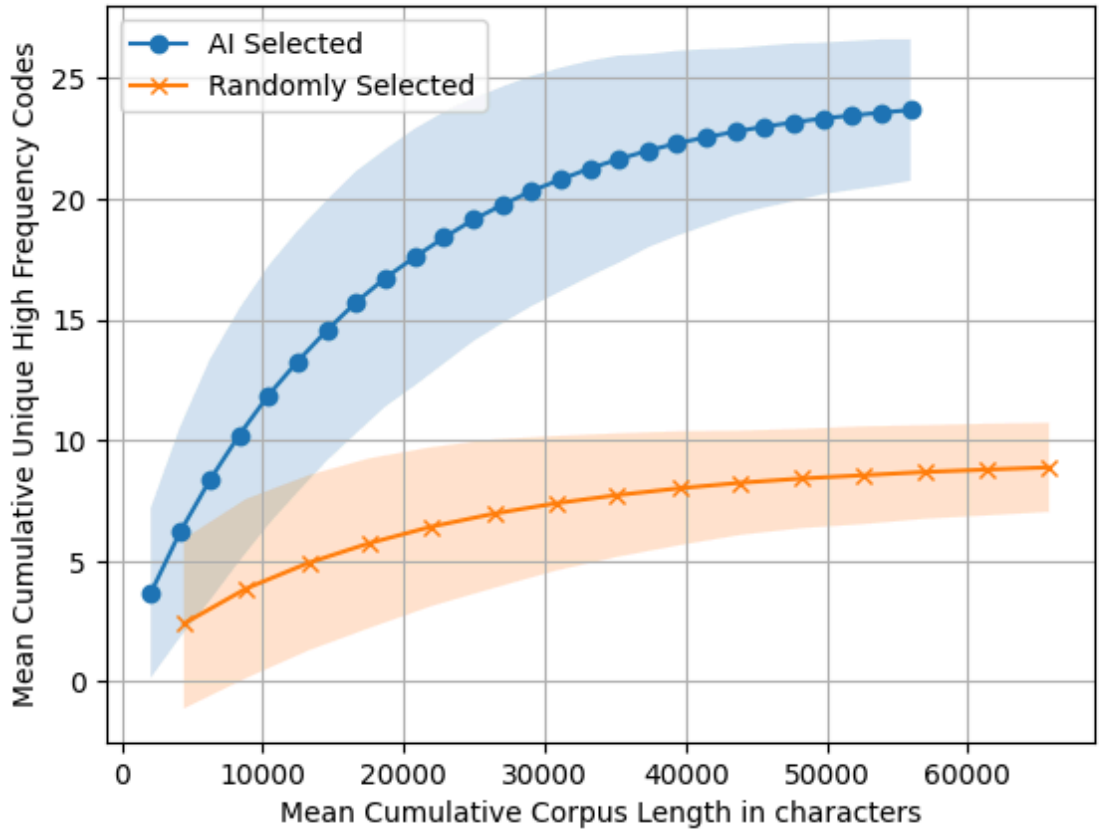

However, there are actually multiple ways to tally cumulative high-frequency codes that yield dramatically different results. Above, we've finalized codebooks, identified high frequency codes, then retrospectively counted how many of these codes have been identified by document i for each cumulative document. We'll term this **retrospective code counting.** Another approach is **iterative code counting**, where we identify new high frequency codes when they first become high frequency–in this case, for the document where a code first achieves 3 instances.

Retrospective code counting is frequently used in papers studying the saturation dynamics of high-frequency codes–though many such papers actually do not specify which method they use. The above results suggest that studies purporting to show quick saturation of high-frequency codes using retrospective code counting are likely to underestimate the number of documents required to achieve their notations of saturation, and will likely find this pseudo-saturation even if no real information redundancy has been achieved.

However, if the lack of new high-frequency codes is to be used as a stopping rule, we can't actually use retrospective code counting, we must use iterative code counting. The same analysis using this method is shown in Figure 9 for the actual results from our document order and Figure 10 for the smoothed results from the bootstrap of random document orders.

In Figure 10 we see that the false saturation induced by retrospective code counting disappears[12]–the cumulative code plots no longer appear to converge to different asymptotes. Indeed, we see no evidence of saturation, even in the AI-selected corpus. Nonetheless, the 10+3 rule is satisfied once for randomly selected and multiple times for AI selected, despite each apparent plateau being followed by numerous new codes, illustrating the limits of such heuristics.

*Figure 9:*

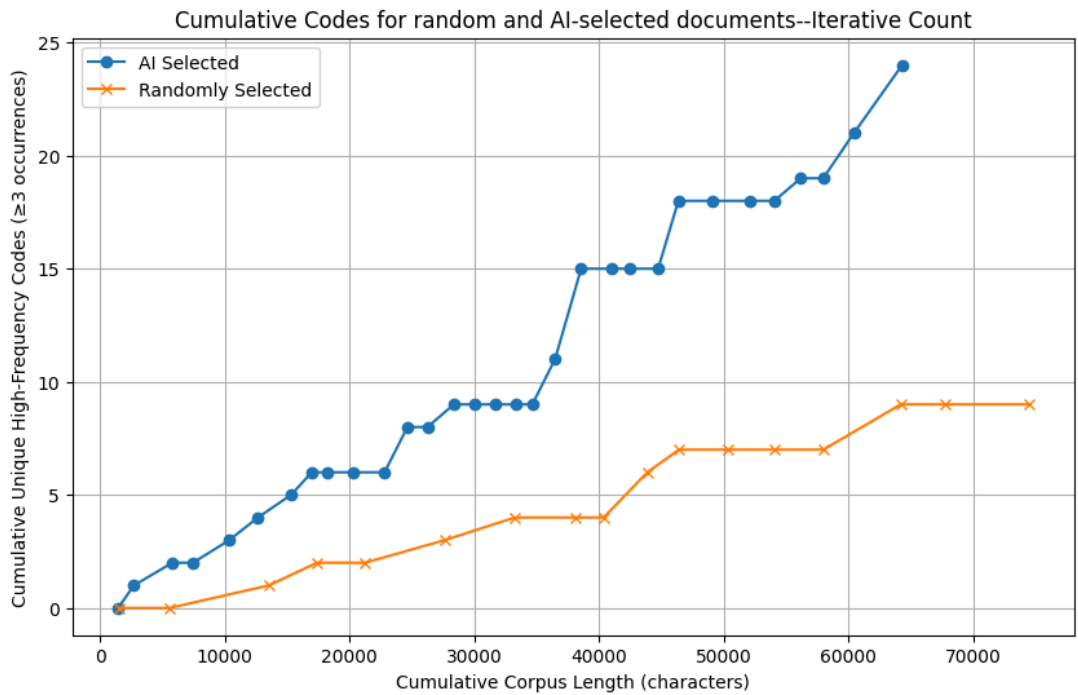

*Figure 10:*

---

[12] We should note that iterative code counting with a fixed threshold for high-frequency codes can itself underestimate the degree of saturation, as a large number of documents are generally necessary to identify a code as high-frequency, even when it's already been recorded in an earlier document.

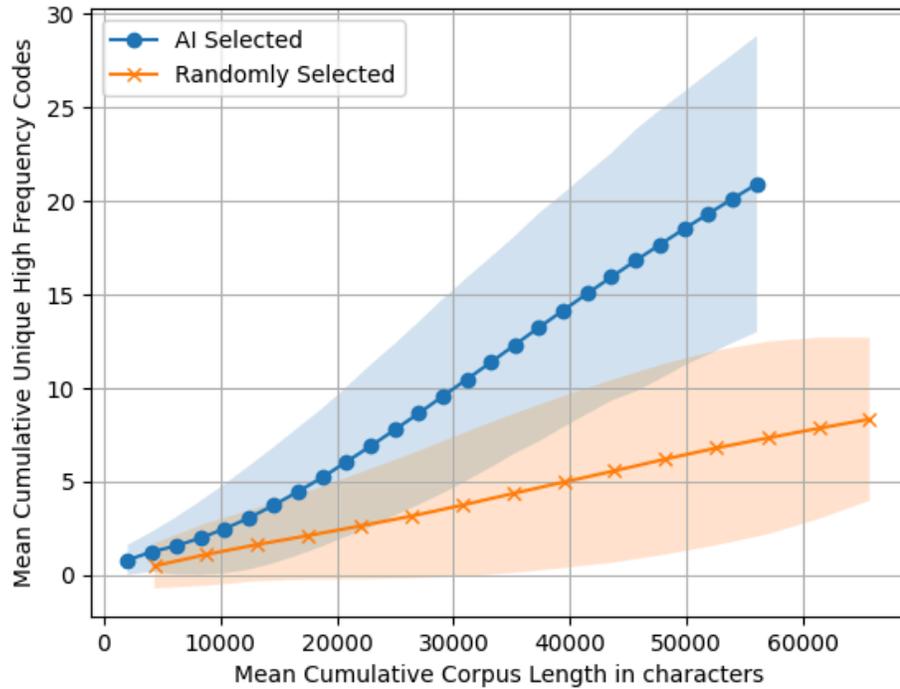

Finally, another common approach is to identify when the set of *themes,* rather than codes, are found (e.g. Hagaman and Wutich, 2017)). We test a version of this method where we count how many unique themes have been discovered by the i*th* document. After completing their coding, we asked our coders to develop a list of themes (need more details here) as well as a mapping from codes to the themes they informed. Figure 11 shows how themes accumulated for AI- and randomly-selected articles: by a corpus size of 30,000 characters codes touching on almost all themes had been discovered, while the random corpus never reaches that level of saturation in the complete corpus of over 70,000 characters. For the first time, we see pronounced flattening of our saturation metric that isn't pathological. Figure 12 smooths the curves by randomizing document order, showing both corpora sharply flattening, but the AI-selected corpus flatting earlier and accumulating themes much faster.

*Figure 11:*

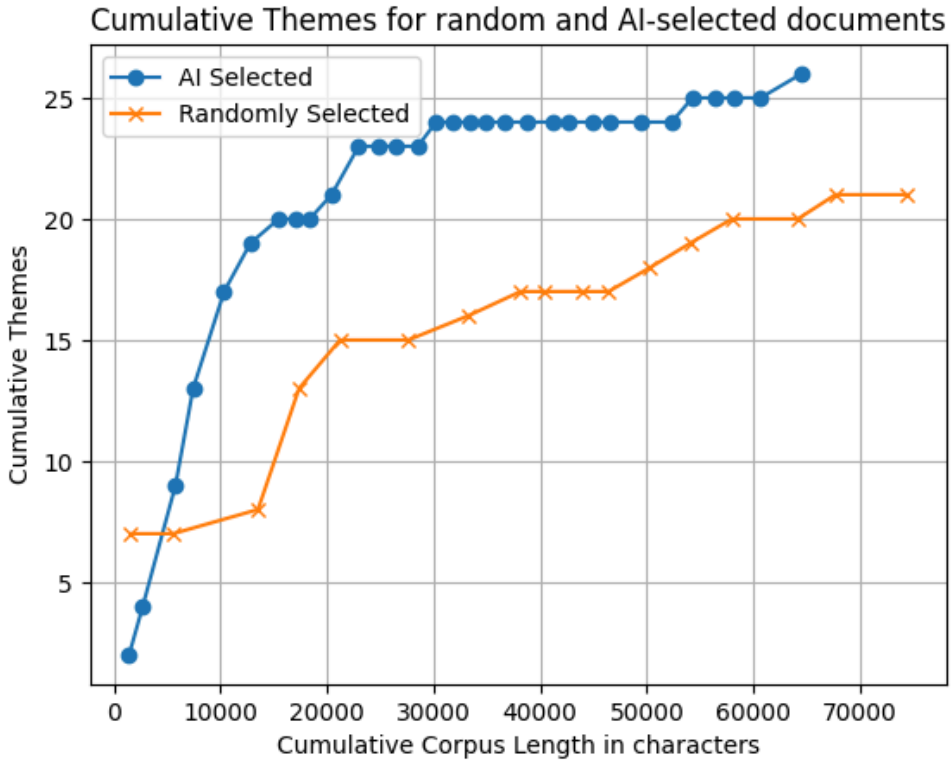

*Figure 12:*

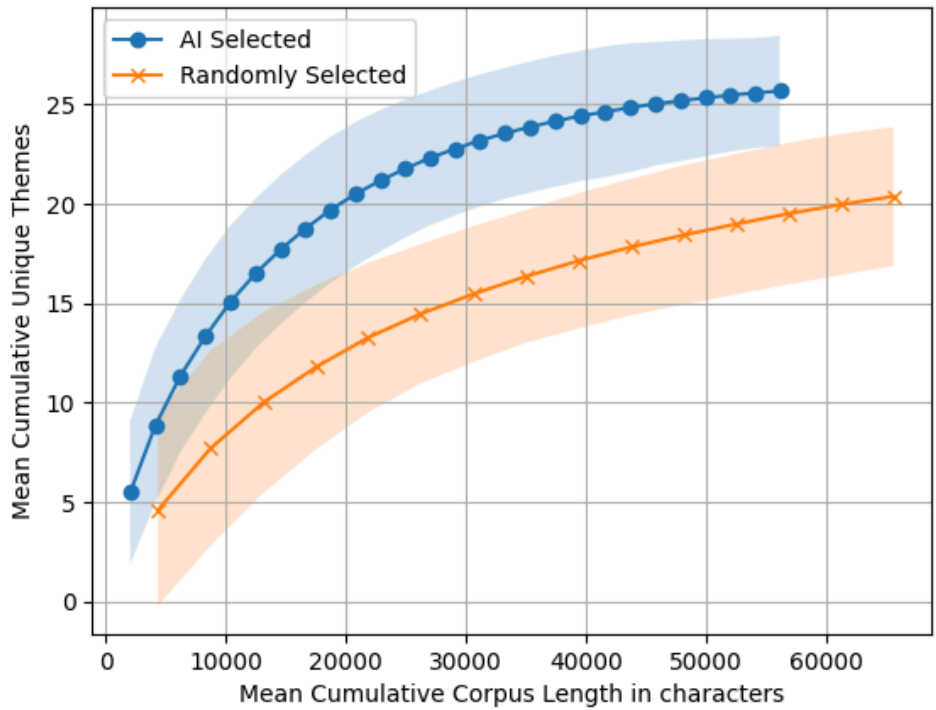

*AI Selection and Documents Superset Size*

We see an impressive increase in fecundity in the AI-selected corpus, but it was selected from a superset of 2530 documents. In many cases collecting and AI-coding so many documents may be infeasible. How useful would this method be with a smaller superset of documents to select from? We lack the resources to do further coding experiments using AI-selected corpora drawn from supersets of varying sizes, but we can do some back-of-the-envelope estimates of what the results might look like.

In this section we generate random subsets of our base superset of 2530 articles. These subsets are of size 50, 100, 250, 500, and 1000. For each size, we generate 10 random subsets of this size and apply our AI-selection method to each for a corpus of length equal to 20 average articles. For each size, we calculate the average predicted fecundity (AI code density) across all 10. Combining this with our results from the full set (2530) and the randomly selected corpus, which is equivalent to selecting 20 out of 20 documents, we have estimates of the average AI code density for selecting from 20 (random selection) as a baseline as well as 50, 100, 250, 500, 1000, and 2530.
We then use our human coded dataset to predict human code density across both rounds by regressing it on AI code density and its square. Then we use the imputed quadratic function to predict human code density for each of these superset sizes based on the AI code density. Figure 13 shows these results, with the predicted human code density normalized to 100% for a randomly selected corpus. These results suggest that selecting documents from a superset of 250 documents provides nearly as much benefit as a superset of 2530 documents, and that about half the benefits are realized when choosing out of just 50 documents.

We must emphasize the limitations of this analysis. The relationship between human code density and AI code density is noisy, not deterministic, and the dataset we use to estimate the relationship is small so the quadratic function is estimated with significant error. Further, the constructs of AI code density and human code density depend on the corpus they are evaluated in, so extrapolating them to counterfactual corpora generated under different conditions may introduce further error. Indeed, the predicted values for 2530 and 20 in Figure 13 differ slightly from the true densities our coders found. However, we believe this is a reasonable, if rough, first attempt at a prediction of the benefits of using AI selection from various superset sizes.

*Figure 13:*

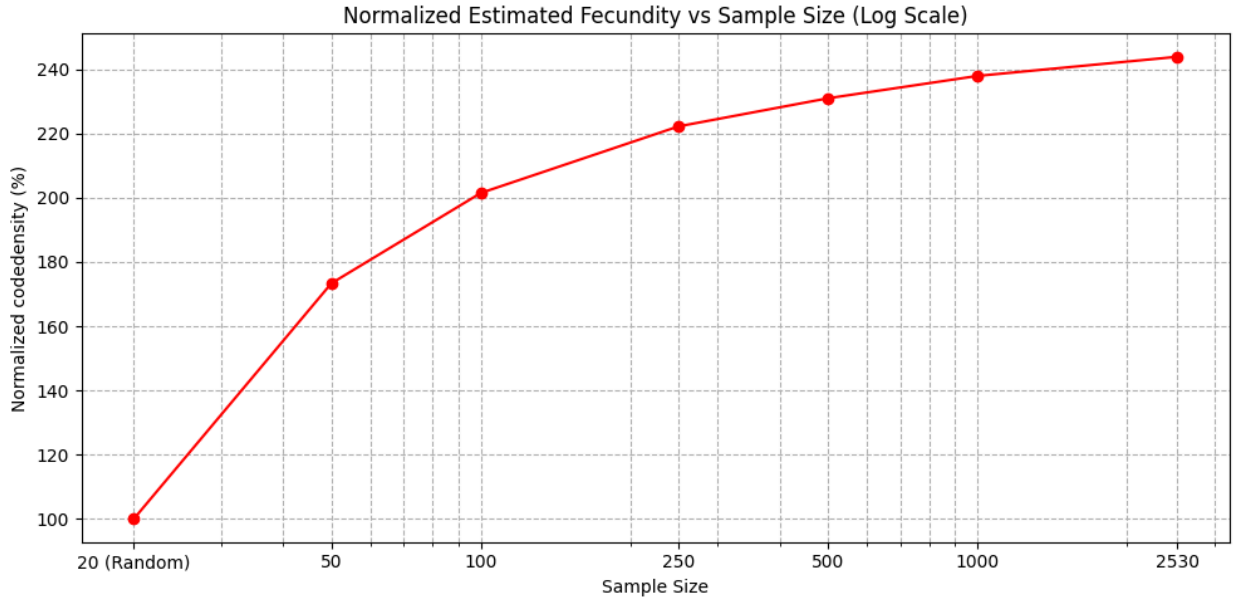

*Discussion*

In this paper we explored how AI-identified fecundity–measured through AI-generated codes–can be used to identify fecund documents–documents denser with text that can be made meaningful in a particular thematic analysis. This not only provides a method to improve corpus selection, decreasing corpus size or increasing the range of issues that can be surfaced in a corpus of a given size, but also sheds light on issues surrounding saturation, where we find that code saturation cannot plausibly be achieved even with the very large corpus we work with, while theme saturation appears a much more reasonable metric. We also find that frequent code saturation can conclude all common themes have been identified when only a fraction of common codes have been identified. Future issues suggested by this research include the question of whether AI generated codes can predict saturation and therefore appropriate corpus size for researchers that wish to use saturation for that purpose, more rigorous analysis of the relationship between superset size and the fecundity of AI-selected documents, and improvements to the AI selection process itself. The AI-generated codes in this paper are generated using a LLM, GPT-3.5-Turbo, that is even at the time of this writing dated. While this method can approximately double the fecundity of the corpus, the highest fecundity documents in our dataset suggest a perfect fecundity predictor might be able to select a corpus four times denser than a randomly selected corpus.

# Appendix A

*Table A1: News Sources Selected for Dataset*

| Source | Included | Notes |
| --- | --- | --- |
| The Star | Yes | Data only available from Jan 2020 |
| Malay Mail | Yes | |
| Malaysiakini | Yes | |
| FMT | No | Scraping not allowed |
| NST | No | Scraping not allowed |
| MalaysiaNow | No | Search query only allows "or" logical operator, not "and" so malaysia refugees generates largely irrelevant results. |
| Yahoo Malaysia | No | Just an index of other news sources |
| Bernama | No | Search results available only for past 2 weeks |
| TheMalaysianInsight | Yes | |
| Al Jazeera | No | Not local |
| The Sun | Yes | |

*Figure A2: Paragraph (linebreak) Counts and Article Lengths for the Full Dataset*

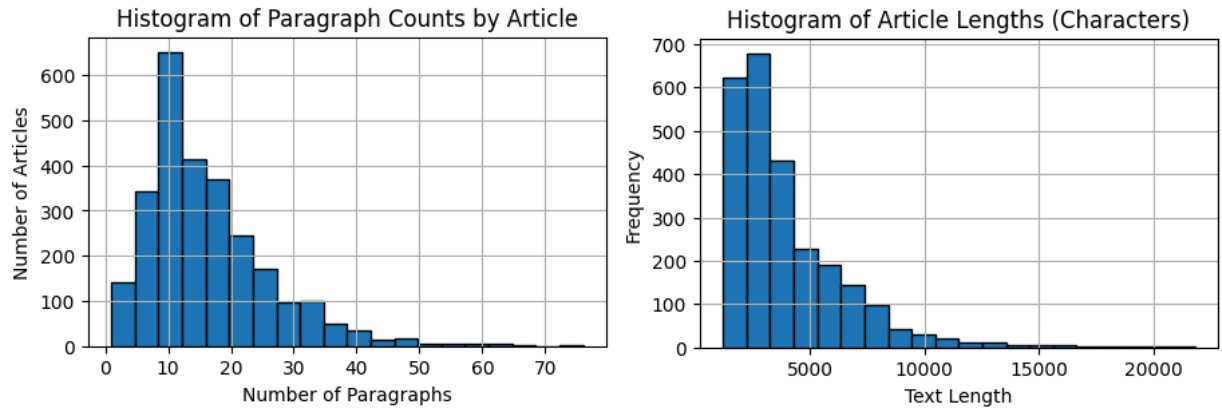

# Appendix B

Below is a summary of our approach to generating Round 2 codes. A full exposition can be found in Flanders, Nungsari, and Cheong (2025)

First, we identified numerous types of irrelevant paragraphs and built a pre-analysis pipeline to exclude them, utilizing OpenAI's Ada-2 embedding, UMAP, and HDBSCAN to cluster the passages by semantic content and manually identifying irrelevant clusters. We increased the the minimum paragraph length threshold from 50 to 100, shortened the article summaries provided to GPT, and switched to a structured JSON prompt breaking the coding process into four distinct components to improve detail in the responses.

In addition to this, we partitioned the full dataset into 400 clusters, sampled four passages from each cluster, and used a more complex, multi-step socratic coding process on those 1,600 passages to generate high quality codes. When generating our final codings on the full dataset, we identified which of the 400 clusters that passage belonged to and provided the 4 high quality codes from that cluster as context to GPT.

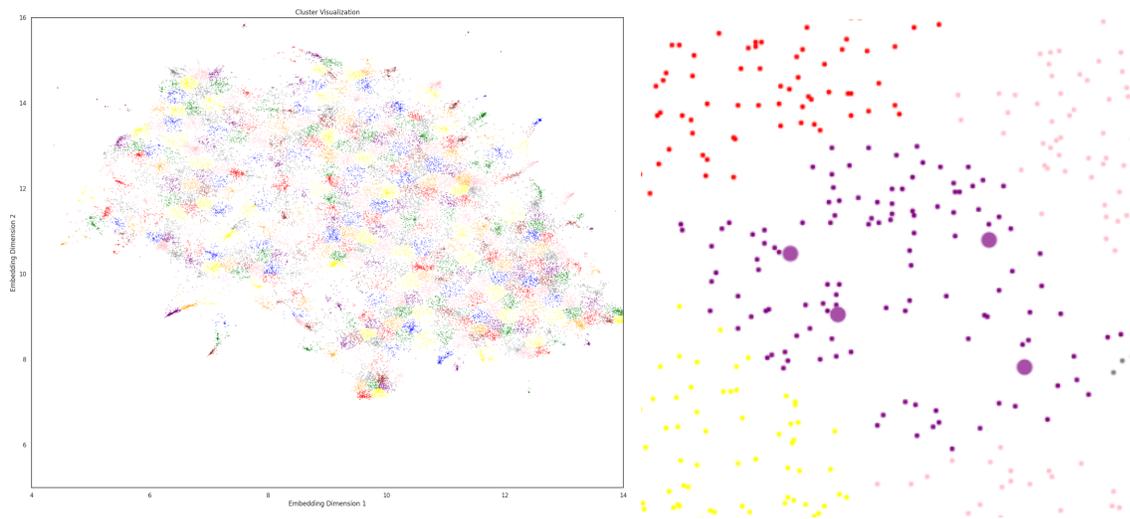

*Figure B1: 2D Visualization of the 400 clusters. B: Detail of one cluster with the four sampled passages highlighted*

The longform Socratic approach was as follows: we asked GPT to assess whether the passages were photo captions or a disclaimers:

```
'Read a passage from a news article ### ' + str(excerpt)+ ' ### Is this passage a piece of text such as
1. a disclaimer of opinion, 2. a photo caption. Or is it a complete passage from the body of a news
article? Respond only in the following python dictionary format: {"1. disclaimer?": True/False, "2.
caption?": True/False, "Body?": True/False }'
```

Then we asked GPT to assess whether the passages explicitly discuss refugees and Malaysia:

```
'Read a passage from a news article ### ' + str(excerpt)+ ' ### Step by step, answer the following
questions: 1. Does the passage explicitly, unambiguously discuss refugees? Note: most passages are not
about refugees. 2. Does the passage explicitly, unambiguously reference Malaysia? Note: most passages
are about other countries.' + note + ' Now respond in the following Python dictionary format: {"1.
Refugees?": "Yes."/"No.", "2. Malaysia?": "Yes."/"No."}'
```

The *note* variable feeds GPT's analysis in the previous step(s) back into the current step. If no red flags were raised, it is empty: "", however, if any red flags were raised, it summarizes the previous issues. In this step it will flag the possibility that "Passage is a disclaimer of personal opinion" or "Passage is a photo caption".

Next, we elicited GPT's confidence that the passage was actually relevant and why not if not:

```
'Read a passage from a news article ### ' + str(excerpt)+ ' ### Answer step by step: 1. Might this
passage be relevant to attitudes towards refugees in Malaysia? If it clearly is, answer "Yes." If it
might be, depending on the context of the article the passage is from--eg. the identity of the subject
and their location--answer "Maybe." If it is definitely irrelevant regardless of context, answer "No."
' + note + ' 2. If "No." or "Maybe.", in 15 words or less give any and all reasons why it might be
irrelevant--both those provided earlier and any others you identify, such as irrelevant output from a
```

```
content management system or editorial annotations to the article. Respond in the following python
dictionary format: {"1. Relevant?":"Yes."/"Maybe."/"No.","2. Why Not?":string or None} '
```

Here, the *note* variable is empty if no red flags were raised: "", however, if any red flags were raised, it adds the following text to the prompt: *'Note: this passage has been flagged as possibly meeting the following criteria for irrelevance:  [CRITERIA] ### If any of these criteria are true, you should answer "No." or "Maybe.  "*, where [CRITERIA]  are given as  *'Passage is a disclaimer of personal opinion, ',  ' Passage is a photo caption, ',' Not about refugees, ', ' Not about Malaysia, '* as relevant.

In the fourth prompt, we ask GPT to code the passage, taking into account its previous assessment of relevance.

```
'Read a passage from a news article ### ' + str(excerpt)+ ' ### Give the theme of this passage as it
embodies, relates to or reflects attitudes towards refugees in Malaysia if it is relevant to that
topic. If it is not relevant, simply summarize the passage in a few words. Note that this passage may
simply be text from the web interface and not from an article at all. Before answering, analyze step by
step: 1. in 14 words or less, return the theme. Do not offer a generic theme like "attitudes towards
refugees in Malaysia", but give a specific theme. 2. Whose attitudes are being reflected? Examples: the
Malaysian government, The Bangladeshi government, Malaysians, NGOs, the author. 3. Who is the target of
the attitudes? Examples: migrant workers, Myanmar, the Rohingya, the government, UNHCR. 4. What is the
valence of attitudes towards the target, if any?:  "Sympathetic.", "Hostile.", or "N/A". ' + note + '
Finally, Respond ONLY in the following python dictionary format: {"1. Theme": stringval1, "2. Whose
Attitude?":stringval2,"3. Target":stringval3,"4. Valence": "Sympathetic."/"Hostile."/"N/A"}'
```

In this prompt, the variable *note* is empty if "Relevant?" was assessed as "Yes" in the previous step, *' Previous analysis found that this passage might be irrelevant for this reason: ' + reason + '### Take this into account.'* if "Relevant?" was "Maybe", and *' Previous analysis found that this passage is irrelevant for this reason: ' + reason + '### Take this into account.'* if "No", where *reason* was the reason given in "Why Not?" in the previous step.

Note that, in the Socratic approach, the initial coding does not reference the summary. Instead, we first see if the passage stands on its own as relevant, then ask GPT to reassess its code in light of the summary. This helps prevent summary bleeding, where GPT's coding is only relevant to the summary, not the passage of interest. This leads us to our final coding step:

```
'Read a passage from a news article ### ' + str(excerpt)+ ' ### The theme of this passage was coded as
### ' + str(precode) + ' ### but this analysis ignores the article summary and is therefore unreliable.
Reassess the theme of this passage as it relates to attitudes towards refugees in Malaysia, given the
context of this summary of the article it came from ### ' + str(summary) + ' ### Before answering,
analyze step by step: 1. in 14 words or less, return the reassessed theme (if relevant) as it relates
to attitudes towards refugees in Malaysia, or return None. Do not give a generic theme like "attitudes
towards refugees in Malaysia", but provide a specific theme. If irrelevant, return None for all further
questions. If relevant, 2. Whose attitudes are being reflected? Examples: the government, Malaysians,
NGOs, the author. 3. Who is the target of the attitudes? Examples: the Rohingya, the government, UNHCR.
4. What is the valence of the attitude towards the target, if any?:  "Sympathetic.", "Hostile.", or
"N/A". ' + note + ' Once again, the passage to code is ### ' +  str(excerpt)+ ' ### Finally, Respond
```

```
ONLY in the following python dictionary format: {"1. Theme": stringval1/None, "2. Whose
Attitude?":stringval2,"3. Target":stringval3,4. Valence": "Sympathetic."/"Hostile."/"N/A"}'
```

In this prompt, *precode* is the code from the previous step and *note* contains *' Previous analysis found that this SPECIFIC passage might be irrelevant for this reason: ' + reason + '### Does the summary clarify this?.'*

Or *' Previous analysis found that this SPECIFIC passage might be irrelevant for this reason: ' + reason + '### Does the summary clarify this?.'*

If relevance was assessed as maybe or no, respectively.

At this point, we grouped each set of 4 codes for each of the 400 clusters and generated a summary for each:

```
'Read a list of four themes from a cluster of passages ### ' + str(codes)+ ' ### Step by step, answer
the following: 1 Are all of these themes both present and relevant to attitudes towards refugees in
Malaysia? "All are."/"None are."/"Some are.". If irrelevant, return none to all further questions.  2.
If relevant, return the overarching theme as it relates to attitudes towards refugees in Malaysia, or
return None. Do not give a generic theme like "attitudes towards refugees in Malaysia", but provide a
specific and detailed theme. If relevant, 3. Whose attitudes are being reflected? Examples: the
government, Malaysians, NGOs, the author. 4. Who is the target of the attitudes? Examples: the
Rohingya, the government, UNHCR. 5. What is the overall valence, if any? Finally, Respond ONLY in the
following python dictionary format: {"1. Are Passages Relevant?":"All are."/"None are."/"Some are.","2.
Theme": stringval1/None, "3. Whose Attitude?":stringval2,"4. Target":stringval3,"5. Valence":
"Sympathetic."/"Hostile."/"N/A"}'
```

Here is the summary for the four codes given earlier:

This leads us to our final prompt, which is run for the entire corpus:

```
'Read this passage from a news article ### ' + str(excerpt) + ' ###  If relevant, give the theme of
this SPECIFIC passage as it embodies, relates to, or reflects attitudes towards refugees in Malaysia.
The following summary of the excerpted article may provide context for the passage (e.g. who is being
discussed and where events are occurring): ### ' + str(summary) + ' ### Here is an overview of how
several passages similar to this one have been coded: ### ' + str(relevant) + ' ### DO NOT copy this
coding verbatim, but use it as reference and be careful if only some or none of the similar passages
were deemed relevant. Before answering, analyze step by step: 1. in 12 words or less, return the theme
(if relevant) as it relates to attitudes towards refugees in Malaysia, or return None. Do not give a
generic theme like "attitudes towards refugees in Malaysia", but provide a specific single theme. If
irrelevant, return None for all further questions. If relevant, 2. Whose attitudes are being reflected?
Examples: the government, Malaysians, NGOs, the author. 3. Who is the target of the attitudes?
Examples: the Rohingya, the government, UNHCR. 4. What is the valence of attitudes towards the target,
if any?:  "Sympathetic.", "Hostile.", or "N/A". Once again, the passage to code is ### ' +
str(excerpt)+ ' ###  Finally, Respond ONLY in the following python dictionary format: {"1. Theme":
None/stringval1, "2. Whose Attitude?":None/stringval2,"3. Target":None/stringval3, 4. Valence":
"Sympathetic."/"Hostile."/"N/A"}'
```

# Appendix C

Bootstrapping is a technique where synthetic samples from a distribution are generated by resampling the observed data. In our case, we sample the documents to be coded, which also randomizes the order in which documents are added to the corpus, and we simulate the cumulative counts for 2,000 such random orderings for each plot. Different documents have different numbers of codes, so the order in which they are tallied will change the shape of the cumulative count curve.

We include not just the average cumulative count, but also a 95% confidence interval. The most common method for bootstrap confidence interval estimation is to use sampling with replacement and simply use e.g. the 2.5th and 97.5th percentile of the simulated results. However, we cannot sample with replacement in this setting. For many statistics like averages, sampling with replacement works fine. However, we are not calculating an average, but a count of unique values. Sampling with replacement means the same document can be sampled more than once, and such a document must contribute 0 marginal unique values as those values were already present in the first copy–it biases the unique value count down. Instead we must use sampling without replacement.

However, we cannot directly calculate confidence intervals from the distribution of outcomes in the presence of a finite population (a finite set of coded documents). To see why, consider the simulated values for the final document to be added. Since that document completes the corpus, the number of unique codes will always be the same, and the computed confidence interval will have width 0. In fact, this dynamic causes the confidence intervals to be underestimated across the cumulative curve, the bias increasing with each cumulative step. To address this, we multiply the empirical 2.5 to 97.5 percentile without-replacement interval by a finite population correction (FPC). We also drop the last 10% of cumulative documents from our bootstrap analysis, as, while the FPC can correct for this bias in expectation, it becomes very noisy when sampling the last few elements of a population. For example, the FPC-adjusted width of the confidence interval for the final document is the indeterminate 0/0.

Finally, because we are graphing unique codes against text length rather than document count, the x-axis position of the i*th* document actually varies across iterations. We take the mean across all 2,000 iterations for each cumulative document, leading to the approximately equal spacing between documents.